\documentclass[conference]{IEEEtran}
\IEEEoverridecommandlockouts
\usepackage{float}
\usepackage{cite}
\usepackage{amsmath,amssymb,amsfonts}
\usepackage{algorithmic}
\usepackage{graphicx}
\usepackage{textcomp}
\usepackage{adjustbox} 
\usepackage{xcolor}
\usepackage{booktabs}
\usepackage{xurl}
\usepackage{flushend}
\usepackage{subcaption}
\usepackage{xurl}

\def\BibTeX{{\rm B\kern-.05em{\sc i\kern-.025em b}\kern-.08em
    T\kern-.1667em\lower.7ex\hbox{E}\kern-.125emX}}

\begin{document}

\title{Representation Bias in Political Sample Simulations with Large Language Models}

\makeatletter
\newcommand{\linebreakand}{%
  \end{@IEEEauthorhalign}
  \hfill\mbox{}\par
  \mbox{}\hfill\begin{@IEEEauthorhalign}
}
\makeatother

\author{\IEEEauthorblockN{Weihong Qi}
\IEEEauthorblockA{\textit{Department of Political Science} \\
\textit{University of Rochester}\\
Rochester, USA \\
wqi3@ur.rochester.edu}
\and

\IEEEauthorblockN{Hanjia Lyu}
\IEEEauthorblockA{\textit{Department of Computer Science} \\
\textit{University of Rochester}\\
Rochester, USA \\
hlyu5@ur.rochester.edu}
\and

\IEEEauthorblockN{Jiebo Luo}
\IEEEauthorblockA{\textit{Department of Computer Science} \\
\textit{University of Rochester}\\
Rochester, USA \\
jluo@cs.rochester.edu}
}

\maketitle

\begin{abstract}
This study seeks to identify and quantify biases in simulating political samples with Large Language Models, specifically focusing on vote choice and public opinion. Using the {\tt GPT-3.5-Turbo} model, we leverage data from the American National Election Studies, German Longitudinal Election Study, Zuobiao Dataset, and China Family Panel Studies to simulate voting behaviors and public opinions. This methodology enables us to examine three types of representation bias: disparities based on the the country's language, demographic groups, and political regime types. The findings reveal that simulation performance is generally better for vote choice than for public opinions, more accurate in English-speaking countries, more effective in bipartisan systems than in multi-partisan systems, and stronger in democratic settings than in authoritarian regimes. These results contribute to enhancing our understanding and developing strategies to mitigate biases in AI applications within the field of computational social science.

\end{abstract}

\begin{IEEEkeywords}
large language model, political science, representation bias
\end{IEEEkeywords}

\section{Introduction}
Generative Language Models (GLMs), particularly Large Language Models (LLMs) represented by the GPT series, have obtained significant interest in political science and broader social science disciplines. The extant body of work highlights the considerable potential of LLMs for both empirical and theoretical applications in these fields~\cite{argyle2023out, horton2023large, ziems2023can, lyu2024llm,mou2024unifying}. In empirical research, proposed uses primarily include simulating human subjects and augmenting datasets~\cite{aher2023using, argyle2023out, dillion2023can}, with claimed advantages such as cost reduction, participant protection, and augmented diversity~\cite{agnew2024illusion}. However, despite these benefits, several challenges persist in the empirical application of LLMs. Central to these concerns are biases towards certain social groups and the limited cognitive capabilities of LLMs. Specifically, \cite{hartmann2023political} uncovers a left-libertarian bias in the ideologically oriented responses of the widely adopted conversational AI, ChatGPT. Meanwhile, \cite{kotek2023gender} uncovers the gender stereotypes embedded within LLM outputs and \cite{agnew2024illusion} argues that employing LLMs in experimental designs could impede efforts towards achieving representative and inclusive samples.

In this study, we concentrate on identifying biases in simulating political samples, particularly focusing on vote choice and public opinion simulations. We emphasize these issues due to the critical nature of interpreting vote choices and understanding public opinions, which are central topics in political science research~\cite{gronke2008convenience, stewart2011voting, berinsky2017measuring}. Furthermore, these areas also represent key directions for applying AI agents to human simulations, underscoring their significance in the advancement of AI-driven methodologies in social sciences~\cite{gao2023s}.

To assess the presence of representation bias in LLM political sample simulations, we first formally define the representation bias as: \textbf{The LLM-based simulations yield significantly better performance for certain groups relative to others.} Moreover, we propose three specific types of representation bias for evaluation:

\begin{itemize}
\item Type 1: Simulation performance is better for samples from English-speaking countries compared to those from non-English speaking countries.
\item Type 2: Simulation performance favors certain demographic groups over other demographic groups (\textit{e.g.,} under-represented ethnic groups versus well-represented ethnic groups).
\item Type 3: Simulation performance varies based on political regimes, with better outcomes for samples from democracies compared to autocracies. 
\end{itemize}

This research primarily leverages {\tt GPT-3.5-Turbo} for the experiments. To identify \textbf{Representation Bias}, following the tradition of human sample simulation tasks~\cite{argyle2023out}, this research intends to use the American National Election Studies (ANES), German Longitudinal Election Study (GLES), Zuobiao Dataset~\cite{pan2018china} and China Family Panel Studies (CFPS) to simulate vote behaviors and public opinions. \textbf{\textit{The simulation performance is evaluated by the agreement proportion of human samples and the LLM-generated results.}}

\section{Related Work}

Representation bias and fairness in AI and LLM development have long been focal points for computer science researchers. Previous studies have concentrated on developing metrics and tools to evaluate AI fairness~\cite{bird2020fairlearn, madaio2022assessing} and on identifying potential biases to enhance the representation of various demographic groups~\cite{ferrara2023fairness, ntoutsi2020bias}. The release of ChatGPT in November 2022 intensified scrutiny on AI applications across various fields, sparking widespread concerns~\cite{qi2024excitements, miyazaki2024public}. This led to a significant surge in research aimed at assessing the fairness of emerging LLMs~\cite{li2023survey}. In general, studies have identified representation biases concerning gender~\cite{wan2023kelly}, culture~\cite{naous2023having}, race~\cite{haim2024s}, and other aspects.

As LLMs continue to evolve, numerous studies have focused on their ability to simulate human samples in surveys and experiments~\cite{aher2023using, argyle2023out}, yet concerns persist about representation biases when LLMs are used as substitutes for human participants~\cite{agnew2024illusion}. While LLM agents demonstrate remarkable capabilities in many scenarios~\cite{xi2023rise,lyu2023gpt}, they also exhibit significant shortcomings in reasoning and other tasks~\cite{lyu2024human, bang2023multitask, zhu2023can, huang2023chatgpt}. However, specific investigations into whether LLMs exhibit representation bias across different political institutions and systems remain limited. This research aims to assess such biases and explore potential sources of bias when LLMs are used to simulate human behavior in political contexts.

\section{Methods}

This study primarily employs {\tt GPT-3.5-Turbo} to assess representation bias in LLM simulations. To quantify representation bias, our study aligns with traditional human sample simulation methods~\cite{argyle2023out}. We leverage datasets such as the American National Election Studies (ANES), the German Longitudinal Election Study (GLES), the Zuobiao Dataset~\cite{pan2018china}, and the China Family Panel Studies (CFPS) to simulate voting behaviors and public opinions.

An example of the designed {\tt prompt} for human sample simulation is as follows:

\begin{center}
\begin{minipage}{.4\textwidth} 
    {\tt System:} Imagine you are a(n) \{{\tt country}\} citizen. \\
    {\tt Prompt:}  You are a \{{\tt race}\} individual and identify as \{{\tt gender}\}. You are \{{\tt age}\}, have \{{\tt highest degree}\}, and your annual income is \{{\tt income}\}. Your religious beliefs align with \{{\tt religious belief}\}. Politically, you identify as \{{\tt ideology}\} and \{{\tt partisanship}\} in party alignment. In the 2020 Election, who will you vote for? Please respond with the name of the candidate.
\end{minipage}
\end{center}

\begin{figure*}[t]
    \centering
    \includegraphics[width=.85\textwidth]{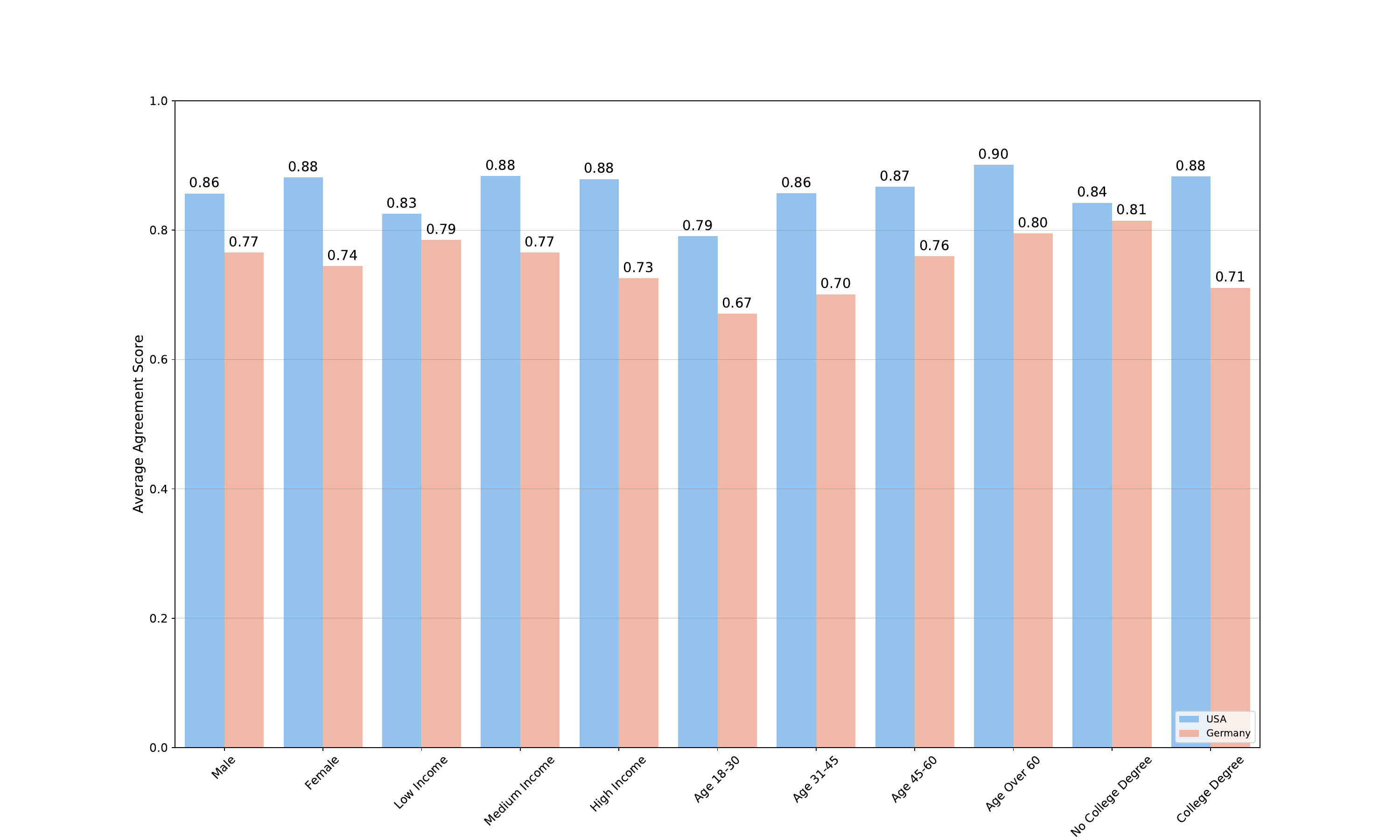}
    \caption{Simulation results for US and German samples in vote choice across demographic groups. A higher agreement score signifies greater similarity between the actual responses from human survey participants and those generated by the simulation.}
    \label{fig: vote_group}
\end{figure*}

Using the ANES data and the GLES data, we can compare the performance of LLM in simulating vote behavior in English-speaking and non-English-speaking countries. Additionally, by comparing  simulation outcomes of public opinions using ANES with those from the two Chinese datasets, we can assess representation bias across various political institutions and different issue dimensions.

To assess the accuracy of LLM simulations of political samples, we adopt the methodology outlined by \cite{aher2023using}, focusing on the agreement percentage between actual responses from human survey participants and those generated by the simulation. Specifically, the agreement score is calculated as follows:
\begin{center}
    $\text{Agreement} \: \text{Score} = \frac{\Sigma_i S_{i, \text{Agree}}}{S_{\text{total}}}$
\end{center}
where $S_{i, \text{Agree}} = 1$ if the actual response and the simulated result are the same for individual $i$ in the sample, and $S_{i, \text{Agree}} = 0$ otherwise. $S_{\text{total}}$ represents the total number of samples in the group.

\section{Results}
\subsection{Vote Choice}

Figure~\ref{fig: vote_group} displays the simulation results from {\tt GPT-3.5-Turbo} for vote choice representation in the USA and Germany, segmented by demographic groups. Generally, simulation performance is superior for the US sample compared to the German sample. This suggests that simulations are more effective in English-speaking, bipartisan electoral systems than in others. Additionally, the results reveal significant disparities in simulation performance across age groups, with notably poorer performance for individuals aged 18 to 30 compared to older demographics. Notably, \textbf{simulation performance improves with the age of the demographic group, indicating challenges in accurately simulating the voting behaviors of younger voters.}

After analyzing the simulation results across countries and demographic groups, we aim to understand the sources of performance disparities. To this end, we dissect the results by partisanship, as illustrated in Figure~\ref{fig: vote_party} which shows that simulation agreement score for both major parties in the USA is high, scoring 0.958 and 0.899 for Democrats and Republicans, respectively. This high accuracy can be attributed to the inclusion of ideological and partisan affiliation data in the ANES, coupled with the tendency of American voters to consistently support their party in elections. In contrast, the simulation results for German major parties reveal a varied pattern. While the results for CDU/CSU, SPD, and AfD are above 0.80, the performance for FDP, GRUENE, and DIE LINKE is significantly poorer.\footnote{CDU/CSU stands for the Christian Democratic Union/Christian Social Union of Germany, SPD represents the Social Democratic Party of Germany, AfD denotes the Alternative for Germany, FDP is short for the Free Democratic Party, GRUENE is the Alliance 90/The Greens, and DIE LINKE refers to the Left.} This variation stems from the complexities of the German electoral system, where coalition formations are common, making it challenging for voter ideology and party membership alone to predict voting behavior accurately. These findings suggest that \textbf{the discrepancies in simulation performance between the US and German samples may arise from the differing voting strategies employed by voters, with the LLM struggling to simulate strategic voting without additional contextual information.}

\begin{figure}[h!]
    \centering
    \includegraphics[width=\linewidth]{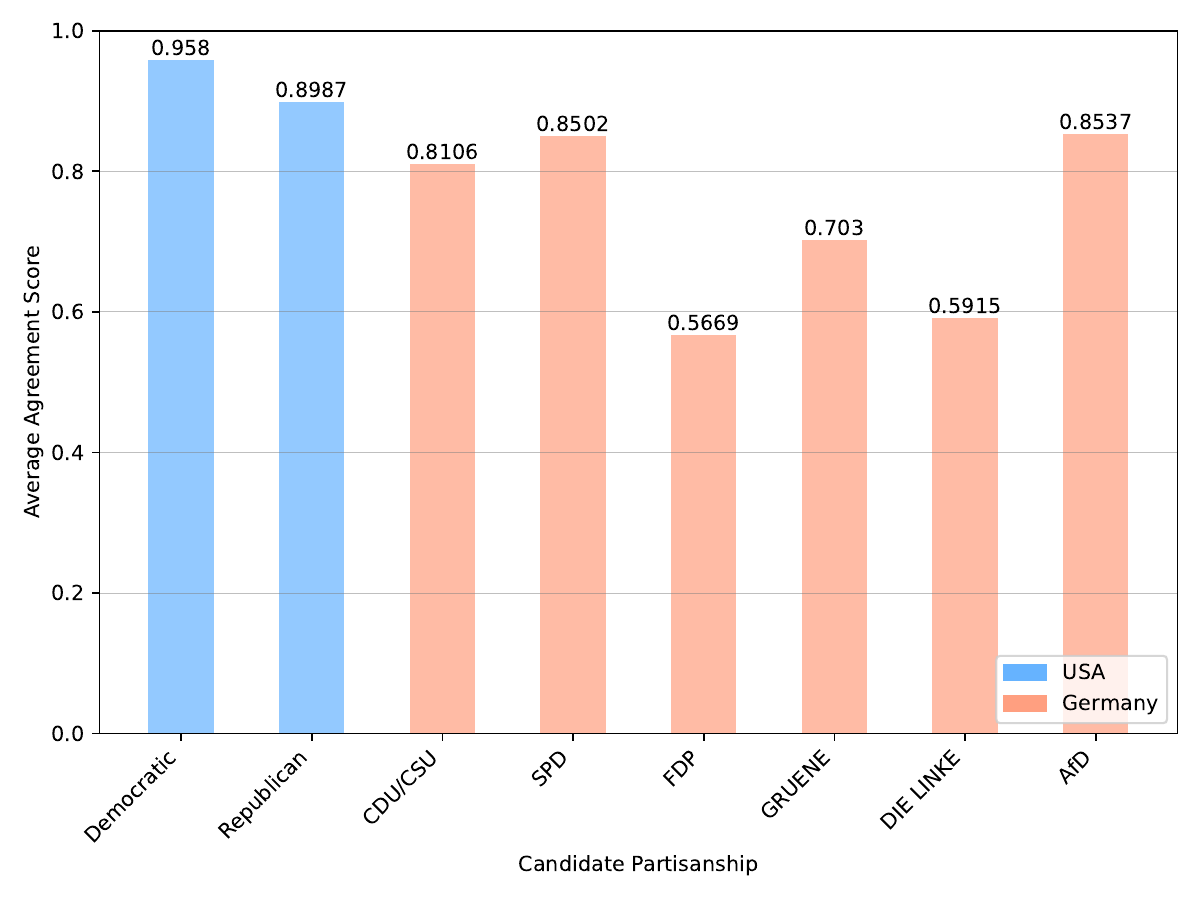}
    \caption{Simulation results for US and German samples in vote choice across parties. A higher agreement score signifies greater similarity between the actual responses from human survey participants and those generated by the simulation.}
    \label{fig: vote_party}
\end{figure}

\begin{figure*}[t]
    \centering
    \includegraphics[width=.7\textwidth]{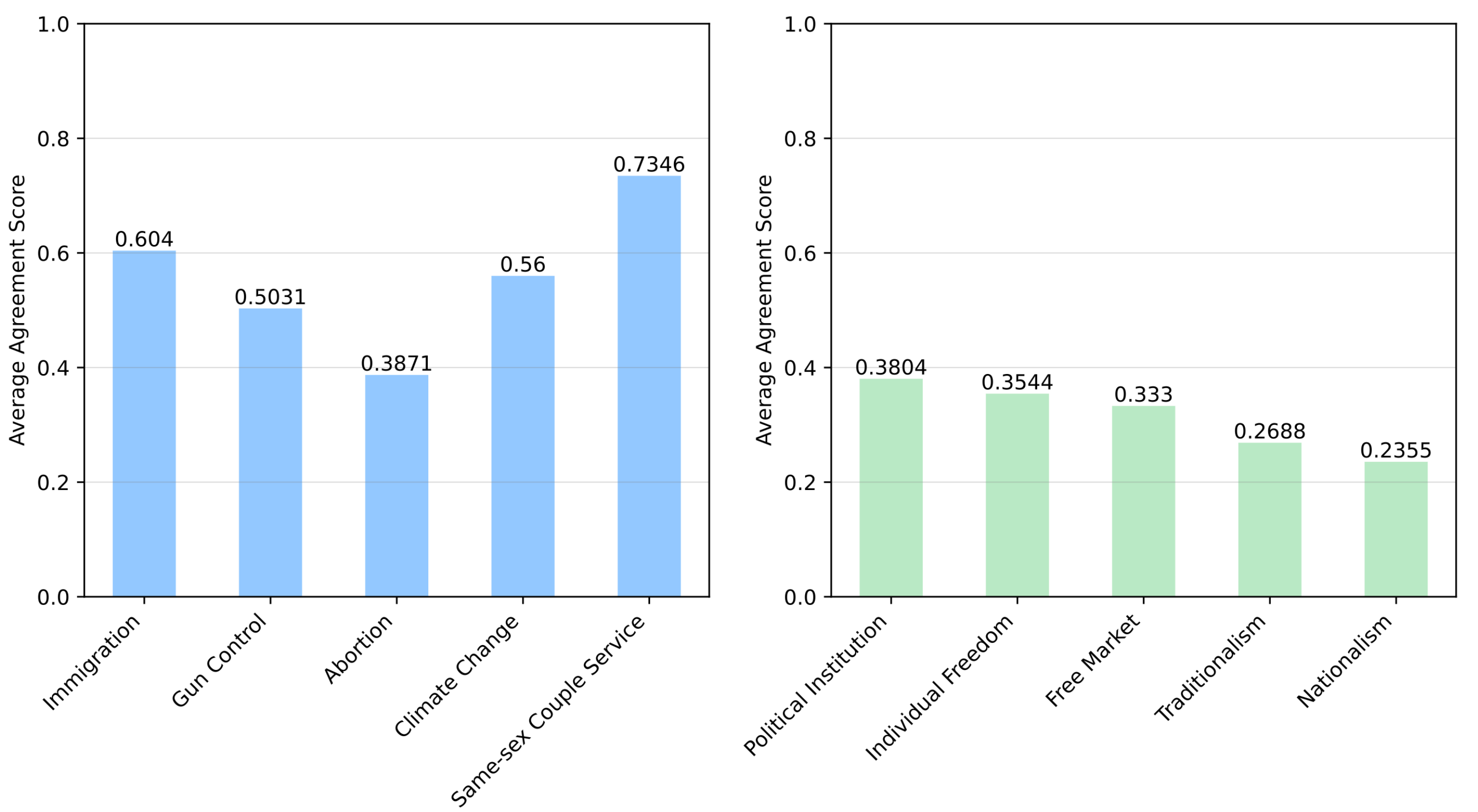}
    \caption{Simulation results for US and Chinese samples in public opinion regarding different political issues. A higher agreement score signifies greater similarity between the actual responses from human survey participants and those generated by the simulation.}
    \label{fig: opinion}
\end{figure*}

\subsection{Public Opinion}

In addition to vote choice, this study also explores the simulation accuracy of GPT regarding public opinions. Unlike vote choice, the salient issue topics vary significantly across different countries and regions. For this evaluation, we employ the top issues identified in the ANES survey and in Chinese ideology research~\cite{pan2018china}. Specifically, for the US sample, we focus on immigration, gun control, abortion, climate change, and services for same-sex couples. For the Chinese sample, the topics under study include political institutions, individual freedom, free market principles, traditionalism, and nationalism.

Figure~\ref{fig: opinion} presents the simulation results of public opinions using the ANES and Zuobiao datasets. Two observations are particularly noteworthy. \textbf{First, simulation accuracy for public opinions is significantly lower than for vote choice}, regardless of languages, countries, or political regimes.  This lower accuracy may stem from public opinions being less predictably influenced by ideology and partisanship alone. Second, \textbf{simulation performance for the US sample is better than that of the Chinese sample, suggesting better representation in English-speaking, democratic contexts.} Specifically, the results for the Chinese sample are notably poor, nearing the level of random guessing. This discrepancy is likely due to the predominance of English in GPT training corpora and the primarily Western, English-speaking background of the model developers.

Based on the simulation results for vote choice and public opinions, we can identify and highlight the following representation biases:

\begin{itemize}
    \item Simulation performance is superior in predicting vote choice but less accurate for public opinions.
    \item Simulation performance excels in English-speaking, bipartisan, and democratic settings, but lags in non-English-speaking, multi-partisan, and authoritarian regimes.
    \item Simulation results are more accurate among older age groups and less reliable for younger demographics.
\end{itemize}

\section{Discussions and Conclusions}
In this study, we investigate the LLM simulation of human samples within a political context, focusing specifically on vote choice and public opinions using survey data from the US, Germany, and China. Employing the {\tt GPT-3.5-Turbo} model, our findings indicate better simulation performance in English-speaking, bipartisan, and democratic countries compared to others. Additionally, the results show a variation in accuracy across different age groups, with older age groups being more accurately represented. The discrepancies in simulation outcomes across various electoral systems may be linked to how well ideology and partisanship predict vote choice or public opinions. Furthermore, disparities across different regimes and languages may stem from the linguistic composition of the LLMs’ training corpora.

Our research carries several implications for fairness in AI and LLM development. The representation bias between English and non-English speaking countries underscores the necessity to diversify the training corpora across various languages. Additionally, the disparity observed across different electoral systems reveals the limitations of LLMs in simulating strategic voting and public opinions, which are less influenced by ideology and partisanship compared to vote choices in bipartisan systems. This indicates a need for further efforts to enhance LLMs' ability to simulate human behavior in more complex scenarios. Moreover, the representation bias across different age groups suggests that additional efforts are required to better understand the behaviors and attitudes of younger populations.

\bibliographystyle{IEEEtran}
\bibliography{IEEEfull}

\end{document}